\relax
\documentclass[letterpaper]{article} 
\usepackage{paper}  
\usepackage{times}  
\usepackage{helvet}  
\usepackage{courier}  
\usepackage{url}  
\usepackage{graphicx}  

\usepackage[update,prepend]{epstopdf}
\usepackage{ragged2e,array,booktabs}
\usepackage{CJKutf8}
\usepackage{multirow}
\usepackage{nameref}
\usepackage{threeparttable}
\usepackage{caption} 
\usepackage{pgfplots}           
\pgfplotsset{width=7cm,compat=1.13}
\usepackage{bm}

\begin{document}
%
\title{A Hierarchical Neural Network for Sequence-to-Sequences Learning}
\author{Si Zuo\textsuperscript{1}, Zhimin Xu\textsuperscript{2}\\
\textsuperscript{1}Aalto University, Finland,  
\textsuperscript{2}SharpSight Limited, Hong Kong\\
\textsuperscript{1}si.zuo@aalto.fi, \textsuperscript{2}zmxu@sharpsight.hk\\
}
\maketitle
\begin{abstract}
In recent years, the sequence-to-sequence learning neural networks with attention mechanism have achieved great progress. However, there are still challenges, especially for Neural Machine Translation (NMT), such as lower translation quality on long sentences. In this paper, we present a hierarchical deep neural network architecture to improve the quality of long sentences translation. The proposed network embeds sequence-to-sequence neural networks into a two-level category hierarchy by following the coarse-to-fine paradigm. Long sentences are input by splitting them into shorter sequences, which can be well processed by the coarse category network as the long distance dependencies for short sentences is able to be handled by network based on sequence-to-sequence neural network. Then they are concatenated and corrected by the fine category network. The experiments shows that our method can achieve superior results with higher BLEU(Bilingual Evaluation Understudy) scores, lower perplexity and better performance in imitating expression style and words usage than the traditional networks.
\end{abstract}

\section{Introduction}
Machine translation (MT) is the power of the machine to "automatically translate a natural language text (source language) into another natural language text (target language)" \cite{Russell1995}. 
There are mainly three methods: Rule-based machine translation(RBMT), Statistical machine translation(SMT), and nowadays Neural machine translation(NMT). The end-to-end neural network machine translation method can directly learn the conversion algorithm between languages by passing data through the encoding network and the decoding network.

From a certain perspective, its degree of automation and intelligence of machine translation are constantly improving, and the quality of it has also been significantly improved. The current state-of-the-art of machine translation technologies can be seen in the results of the Workshop on Machine Translation(WMT), one of the most authoritative competition in machine translation held annually from 2006.


Although the BLEU scores of machine translation system evaluations are increasing year by year, compared to the translation results of professional translators, machine translation still has a long way to go due to the puniness the and flexibility of human languages. Besides, for better performance on translation, there is a sentences length limitation while translating due to “Gradient Explosion/Disappearance” problem\cite{Pascanu13}. Even the introduction of Sequence-to-Sequence learning with the help of encoder-decoder framework\cite{Sutskever14}\cite{Cho14} and Long Short-Term memory (LSTM)\cite{Hochreiter97} can just ease the problem. However, there is a need for optimal long sentences translation for some special documents. Such as financial statements, at least from Chinese to English or English to Chinese, there are large number of long sentences and are inevitable. An example of long sentences in English version of financial statements from a HongKong company is: 
\begin{quote}
    Because of the significance of the matters described in the “Basis for Disclaimer of Opinion” section of our report , we have not been able to obtain sufficient appropriate audit evidence to provide a basis for an audit opinion on these consolidated financial statements and whether the consolidated financial statements have been properly prepared in compliance with the disclosure requirements of the Hong Kong Companies Ordinance .
\end{quote}

Consequently, in this paper, a hierarchical neural network targeting on long sentences translation as well as a dataset from financial statements will be proposed.

\section{Related Work}
Before the development of NMT, Statistical Machine Translation (SMT) dominated this field for a long time. With the development of statistics, researchers applied statistical models to machine translation. This method is based on the analysis of bilingual text corpora to generate translation results. 

In 1997, A idea of using an “encoder-decoder” structure for machine translation \cite{Neco97} is proposed. A few years later, in 2003, a new proposed neural network-based language model\cite{Bengio13} improved the data sparsity of traditional SMT models. Their research lays the foundation for the future application of neural networks in machine translation.

In 2013, a new end-to-end encoder-decoder architecture for machine translation\cite{Kalchbrenner13} is introduced. It uses Convolutional Neural Network (CNN) to encode a given piece of source text into a continuous vector, and then use a Recurrent Neural Network (RNN) as a decoder to transform the state vector into a target language, which can be said to be the birth of NMT. 

The NMT nonlinear mapping between natural languages is different from the linear SMT model, and it uses the state vector connecting the encoder and the decoder to describe the semantic equivalent relationship. However, the “Gradient Explosion/Disappearance” problem\cite{Pascanu13} makes RNN practically difficult to handle long distance dependencies; therefore, the NMT model initially performed poorly. 

One year later in 2014,  a method called Sequence-to-Sequence learning with encoders and decoders\cite{Sutskever14}\cite{Cho14} as well as long short-term memory (LSTM)\cite{Hochreiter97} are introduced for NMT. With the help of the gate mechanism, the “gradient explosion/disappearance” problem is controlled so that the model can obtain far longer sentences “long-distance dependence”.

At the same time the problem of NMT turns into a "fixed-length vector" problem: regardless of the length of the source sentence, it will be compressed into a  fixed-length vector by this neural network needs, bringing more complexity and uncertainty in the decoding process, especially when the source sentence is very long\cite{Cho14}.

Since 2014, "attention mechanism" for NMT\cite{Bahdanau14} is introduced for solving the "fixed length vector" problem. When the decoder generates a word for constructing a target sentence, only a small portion of the source sentence is relevant; therefore, a content-based attention mechanism can be applied to dynamically generate a weighted context vector based on the source sentence and the network then predicts words based on this context vector rather than a fixed-length vector. Since then, the performance of NMT has been significantly improved. Encoder-Decoder Neural Network with Attention Mechanism has become the best model in the NMT field and given the state-of-art performance.

In the meanwhile, there are also other network structures for machine translation. In 2017, Facebook's Artificial Intelligence Research Institute (FAIR) announced that they use CNN to solve translation problem, which can achieve performance similar to RNN-based NMT\cite{Gehring16,Gehring17}, but at a speed that is 9 times faster. In response, Google released a completely new model Transformer which only based on attention mechanism\cite{DBLP:journals/corr/VaswaniSPUJGKP17} in June. This model neither uses CNN nor uses RNN, but is entirely based on the attention mechanism. Inspiring by Generative Adversarial Networks(GANs)\cite{Goodfellow:2014:GAN:2969033.2969125}, for the first time, the method of generative adversarial learning was introduced into the field of machine translation, and a new machine translation learning model based on generative adversarial learning and deep reinforcement learning was proposed.

Due to the limitation of parallel data, making use of monolingual data can boost translation performance. Recently, researchers are exploring unsupervised methods for machine translation\cite{DBLP:journals/corr/abs-1710-11041,DBLP:journals/corr/abs-1711-00043}, which relay on the combination of parallel data and monolingual data or only relay on monolingual data. Keys for unsupervised learning is, firstly initializing model with bilingual dictionary. Secondly, establishing a denoising self-encoder\cite{Vincent:2008:ECR:1390156.1390294} to learn useful information from input data. Finally, via back-translation\cite{DBLP:journals/corr/SennrichHB15a}, generates sentences pairs and turns unsupervised learning into supervised learning.

\section{Proposed Network}\label{sec:proposed_N}
Even though there are different models applied for machine translation, still less of them will focus on improving the quality of long sentences translation. In this paper, a machine translation neural network based on sequence-to-sequence neural machine translation will be proposed, targeting on ease the difficulty of long sentences translation. In the mean while, as we found that in documents like financial statement including large number of long sentences, a dataset extracted from financial statements in both English and Chinese version is prepared. Unlike daily spoken Chinese or written Chinese,  the data from financial statements expresses in a ancient way which means the network will also need to have the ability to imitate words usages and expression style for better translation.

The proposed network consists of a coarse category network and a fine category network. The main idea of the proposed network is learning long sentences hierarchically. The coarse category network level with the aim to provide a better initial state while the fine category network, which can also be considered as a Chinese-to-Chinese coarse category network, takes concatenated output of coarse category network as input and learn semantic compositionality and correct errors at sentences level.

\subsection{Data Segmentation}\label{sec:sec_data_seg}
Before training both networks, all the long sentences are 'cleaned' and segmented into short sequences by rule-based segmentation. Through rough segmentation, the coarse category network can provide a better initial state for long sentences translation. For every part in English, a corresponding part in Chinese which is exactly expressing the same meaning can be found, even though orders might not be the same and conjunctions might be different.

Inspired by the similarities, segmentation is done by combination of words and punctuation according to the frequency of occurrence in both languages. Words consist of relative pronoun and conjunctions. For sentences that cannot segment by words or punctuation or the combination of them, will simply segment by number of words. For example, a sentences with length 90(90 words) will be split into three parts and each part contains 30 words. Such kinds of segmentation might result in wrong mapping between source language and the target language, but experiences results shows that the network has a certain degree of tolerance for such errors. The selected words as well as punctuation are shown in Table \ref{tab:seg_comb} and an example result of segmentation based on combination of words and punctuation for English is shown in Table \ref{tab:seg_punc_word}.

\begin{table}[htb]
\caption{The selected words and punctuation for segmentation}
\label{tab:seg_comb}
\begin{center}
\fbox{
\begin{tabular}{lp{0.7\linewidth}}

{\textbf{English}}  &  '\textbf{,}', '\textbf{which}', '\textbf{and}', 
                       '\textbf{that}', '\textbf{but}' , '\textbf{or}', 
                       '\textbf{so}'\\
\midrule
\multirow{6}*{\textbf{Chinese}} 
    &   '\textbf{\begin{CJK*}{UTF8}{bkai}，(comma)\end{CJK*}}',
        '\textbf{\begin{CJK*}{UTF8}{bkai}、(dun hao)\end{CJK*}}', \\
        
    &   '\textbf{\begin{CJK*}{UTF8}{bkai}和(he)\end{CJK*}}',
        '\textbf{\begin{CJK*}{UTF8}{bkai}并(bing)/并且(bing qie)\end{CJK*}}',\\
    
    &   '\textbf{\begin{CJK*}{UTF8}{bkai}及(ji)/以及(yi ji)\end{CJK*}}', 
        '\textbf{\begin{CJK*}{UTF8}{bkai}其中(qi zhong)\end{CJK*}}',\\
    
    &   '\textbf{\begin{CJK*}{UTF8}{bkai}但(dan)/但是(dan shi)\end{CJK*}}',\\
    &   '\textbf{\begin{CJK*}{UTF8}{bkai}或(huo)/否则(fou ze)\end{CJK*}}', \\
        
    &   '\textbf{\begin{CJK*}{UTF8}{bkai}因此(yin ci)/所以(suo yi)\end{CJK*}}',\\
\end{tabular}}
\end{center}
\end{table}

\begin{table*}[htb]
\caption{Examples of segmentation results using both words and punctuation}\label{tab:seg_punc_word}
\begin{center}
\fbox{
\begin{tabular}{lp{0.8\linewidth}}
\textbf{Before} &  Because of the significance of the matters described in the "Basis for Disclaimer of Opinion" section of our report , we have not been able to obtain sufficient appropriate audit evidence to provide a basis for an audit opinion on these consolidated financial statements and whether the consolidated financial statements have been properly prepared in compliance with the disclosure requirements of the Hong Kong Companies Ordinance . \\
\midrule
\multirow{3}*{\textbf{after}} 
& 1. Because of the significance of the matters described in the “ Basis for Disclaimer of Opinion ” section of our report \textbf{,} \\
& 2. we have not been able to obtain sufficient appropriate audit evidence to provide a basis for an audit opinion on these consolidated financial statements \textbf{and} \\
& 3. whether the consolidated financial statements have been properly prepared in compliance with the disclosure requirements of the Hong Kong Companies Ordinance.
\end{tabular}}
\end{center}
\end{table*}

\subsection{Network Structure}\label{sub_network_struc}
\begin{figure*}[htb]
\centering
\includegraphics[height=8cm]{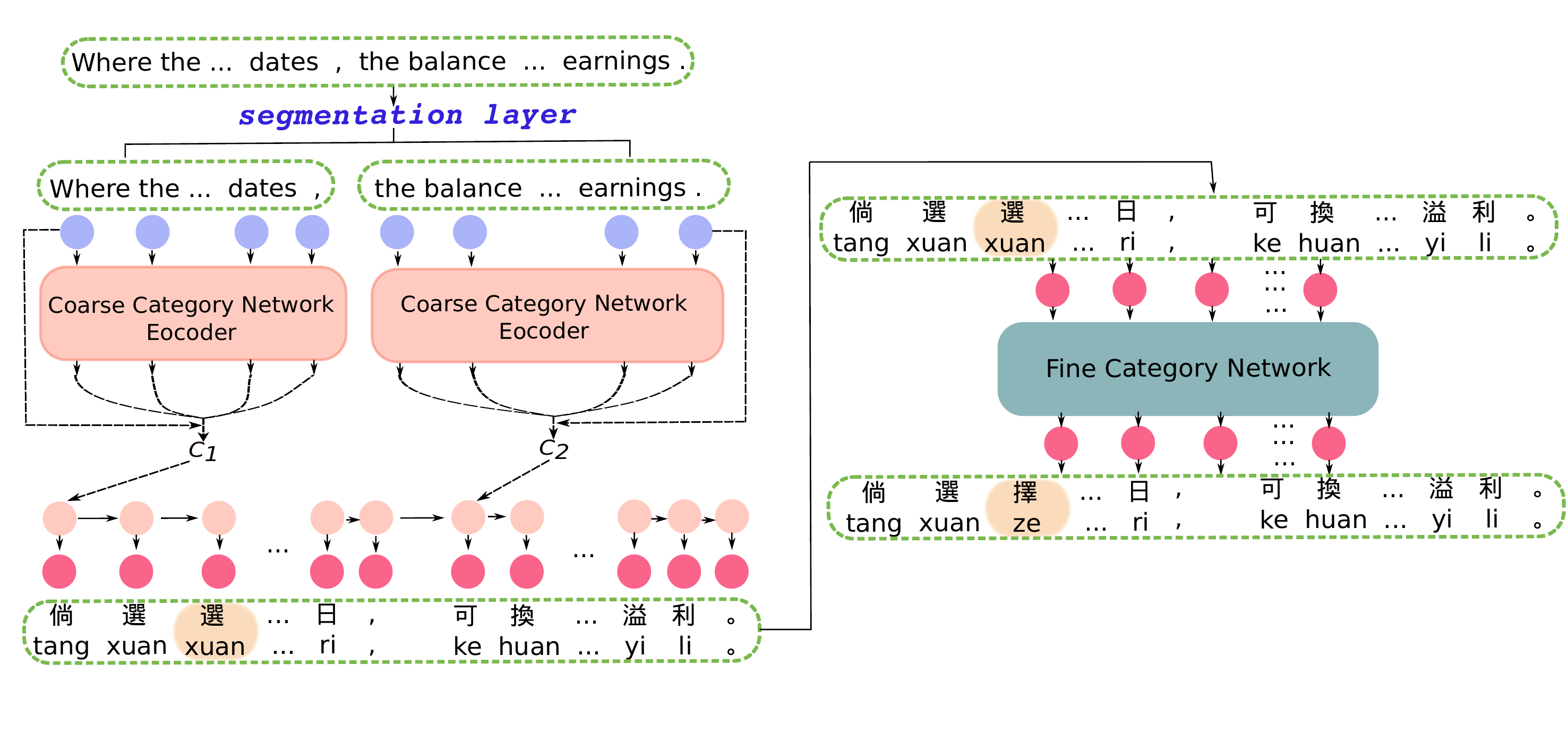}
\caption{Structure of the Proposed Network. The left side of figure represents the coarse category network while the right side shows the fine category network. $c_1$ and $c_2$ in the coarse category network denote the context vector of input short sequences. The output of coarse category network is concatenated and feed into the fine category network.}\label{fig:propsed}
\end{figure*}

The structure of proposed network is shown in Figure \ref{fig:propsed}. The left side of Figure \ref{fig:propsed} represents the word-based English to Chinese coarse category network while the right side shows the Chinese fine category network. $c_1$ and $c_2$ in the coarse category network represents the context vector of input short sequences. The output of coarse category network is concatenated into the original long sentences and directly feed into the fine category network as input. 

\subsubsection{Coarse Category Network}\label{translation_N}
Focus on the coarse category network, after segmentation layer, all the long sentences pairs whose length are larger then 50 will be split into short sequences pairs based on segmentation rules they satisfied. The coarse category network is trained only by sentences whose lengths are less than 50. It consists of two parts: short sentences and short sentences from segmented long sentences. Let the \bm{\mathrm{x}} be the input of long sentence and segmented into $\bm{\mathbf{I}}$ sequences, $\bm{\mathbf{x}=(\mathbf{sx_1}, \mathbf{sx_2},...,\mathbf{sx_I})}$. For each $\bm{\mathbf{sx_i}}$, it contains $j$ words which is less than 50, $\bm{\mathbf{sx_i}=(sx_{i1},sx_{i2},...,sx_{ij})}$.

\subsubsection*{Encoder} 
\begin{figure}[h]
\centering
\includegraphics[height=7cm]{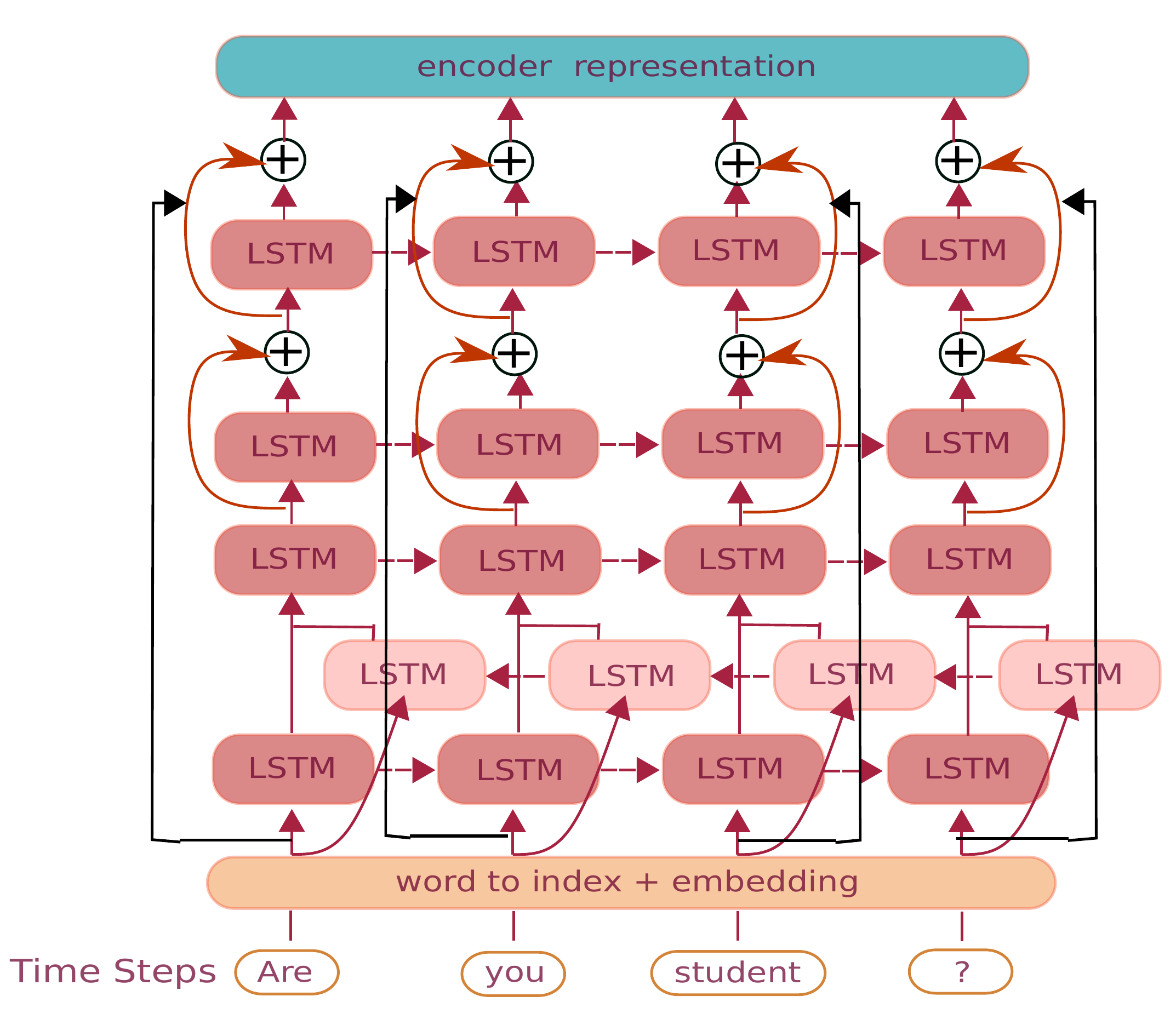}
\caption{Structure of encoder in coarse category network \label{fig:encode_propsed}}
\end{figure}

The structure of encoder in coarse category network is shown in Figure \ref{fig:encode_propsed}. The encoder of coarse category network is a 4-layer LSTM network with 128 hidden units for each LSTM cell. For better learning the global context information and the structure of input data, bidirectional RNNs\cite{Schuster:1997:BRN:2198065.2205129} is introduced into the first layer of encoder, allowing network to learn both backward and forward information about each sequence at every time step:  
\begin{equation}
    \vec h^s_{i,j} = LSTM^s_f(u(sx_{ij}) + \vec h^s_{i,j-1} + b_{ij}) 
\end{equation}
\begin{equation}
    \overleftarrow h^s_{i,j} = LSTM^s_b(u(sx_{ij}) + \overleftarrow h^s_{i,j+1} + b_{ij}) 
\end{equation}
where $LSTM^s_f$ represents forward cell and $LSTM^s_b$ represents the backward cell. And the representation of $j$-th word will be the concatenated output of hidden state of two direction, i.e., $h^s_{i,j}=[(\vec h^s_{i,j})^T, (\overleftarrow h^s_{i,j})^T]^T$.

For better performance, residual connections are added to last two layer of encoder network. As network depth increasing, accuracy of network will be unsurprising saturated and then degrades rapidly. The introduction of residual connection to neural network was aiming to ease gradient vanishing problem and train deeper network, enabling networks to be substantially deeper\cite{He2016DeepRL}. Also, residual connection will provide additional information about previous layer for later layer so that the network can learn more. As a 4-layer LSTM which is actually not that deep, only for the last two layers, a skip connection is added between input of LSTM block and the output of LSTM block. Besides,  to enable the encoder to better represent the complex composition of whole sentences as well as lexical semantic, the embedding input is concatenated with the output of last layer. So, for the third layer of each time step:
\begin{equation}
    \vec h^s_{i,j} = \vec h^s_{i,j}\bigoplus \vec h^s_{i,j_{p\_layer}}
\end{equation}
where $\vec h^s_{i,j_{p\_layer}}$ represents the hidden state of previously layer. After element-wise adding of each component of block input and output, the new output will be fed into next LSTM block. And for the final layer of each time step:
\begin{equation}
     \vec h^s_{i,j} = [\vec h^s_{i,j}\bigoplus \vec h^s_{i,j_{p\_layer}},   \mathbf{\vec sx_i^e}]
\end{equation}
where $\mathbf{\vec sx_i^e}$ denotes the embedding input of sequence $i$. In our case, embedding input is also a part of encoder output, which can be seen from Figure \ref{fig:encode_propsed}. The embedding input provides supplementary information for encoder output, which means for the weighted computing of implementation of attention mechanism in decoder will be more precise as additional information is given.

\subsubsection*{Decoder} The decoder of coarse category network is also a 4-layer LSTM with 128 hidden units. Unlike encoder, the decoder applied neither bidirectional LSTMs nor residual connection, it works together with attention mechanism so as to establish a direct short connection between target and source through the “focus” of the relevant source content during the translation process. The attention mechanism provides a way to not discard the hidden state in the middle of the encoder LSTM, but to allow the decoder to look back at them and figure out which part should focus on at each time step. 

In our decoder, one modification related to attention mechanism is the 'memory' keep for it. The 'memory' input to the decoder including two parts: encoder output from LSTM cells and embedding input as mentioned in Section \nameref{translation_N} for more precisely 'focus' with more information about the source data. Normally, before decoding, the decoder will first calculate the weights between state of current step of the decoder and each step state of the encoder. The decoder generates the translation $\bm{\mathbf{y}}$ with the nonlinear function $f(*):p(y_{k})=f(y_{k-1}, s_{k},a_k)$, where $s_{k}$ is the decoding state while $a_k$ denotes the weighted sum of source context of decoding step $k$. $s_k$ is computed by $g(*): s_k=g(s_{k-1}, y_{k-1}, a_j)$ while $a_k$ is computed by:
\begin{equation}
    a_k = {\sum_{j=1}^{q}{\frac{\mathbf{exp}(d_{k-ij})}{\sum_{j'=1}^{q}\mathbf{exp}(d_{k-i'j})}}}*h_{ij}
\end{equation}
\begin{equation}
    d_{k-ij}=p_{a}^T \mathbf{tan} h(V_a s_{k-1} + W_a h_{ij} + U_a \mathbf{sx_e^i})
\end{equation}
where $d_{k-ij}$ computed based on attention mechanism tells how much does $h_{ij}$ related to $y_k$. 

The decoder sequentially translates segmented short sequences with context vector from encoder. After predicting all the short sequences per sentences, the decoder will output the concatenated result of full long sentences and a basic translation result from English to Chinese is produced.

\subsubsection{Fine Category Network}\label{sec:Correction_N}
After coarse category network, a basic translation result from English to Chinese is produced. As coarse category network learns semantic compositionality at short sequences level, the fine category network, taking concatenated output of translation model as input(which is the original long sentences), learns semantic compositionality and correct errors on output of coarse category network at sentences level.
The structure of fine category network is almost the same as coarse category network except both source and target languages are Chinese, which is shown in the right side of Figure \ref{fig:propsed}. The fine category network is also a Sequence-to-Sequence Neural Network with a encoder and a decoder with attention mechanism, both encoder and decoder are 4-layer LSTM network with 128 hidden units.

\section{Experimental Results}
In this section, conducted experiments for evaluating the performance of proposed methods will be described and the results are also shown. Before going to the experiments, statistics of data set will be introduced. Then results will compared with the predicting results from GNMT\cite{GNMT} and Transformer\cite{DBLP:journals/corr/VaswaniSPUJGKP17} among long test data, short test data and all the test data so as to figure out advantages and weakness of networks.

\subsection{Dataset}\label{sec:data_statics}
\subsubsection{Proposed Data}
Our proposed data is extracted from financial statements from different companies with characters in traditional Chinese. The raw data given is the financial statements in English version with corresponding Chinese. As a financial may contain quite a lot tables and numeral data, paragraph with sentences need to be extracted from the documents and split paragraphs into sentences by hand. One important reason it can not simply split the paragraph automatically for example by symbols(like ',(comma)' or '.(full stop)') is we may fail to get corresponding English and Chinese sentences as sentences segmentation and expression are different in English and Chinese.  After processing paragraphs, each documents provided around 800 useful sentences.

The original number of training data from professional financial statements is 3057. Before training on our own data, we need to first 'clean' the corpus. As the total training size of our own data is relatively small, the quality of data will be significantly important. After filtering, we have 2923 parallel training data with 5074 English words and 1447 Chinese words. Additionally, 100 parallel data pairs are processed as development set and 200 parallel data pairs for test set. We filter the bilingual corpus according to the following criteria:

\begin{itemize}
\item Empty lines and redundant space characters will be removed.
\item Sentences contain illegal characters (such as URLs, characters of other languages) will be removed.
\item Chinese sentences without any Chinese characters will be removed.
\end{itemize}

Besides, as the proposed data from financial statements has only 2923 parallel data pairs for training, it is relatively small and area-specified, which means vocabularies and sentences expressions are limited. One main problem of the proposed data is, same source data may be translated into different target data because of the variety of Chinese expressions and vocabularies. An example of different translation of vocabularies in proposed data is shown in Table \ref{tab:multi_ref}). In this case, when evaluating the performance of proposed network, it may worse the score. It might not be a problem of larger data. However, when it comes to small data, the impact could be significantly obvious. Consequently, data with multi-references will be figured out at the beginning and when evaluating translation performance of those data, predicting result will compare with several reference and the prediction with highest BLEU score will be selected as the final output.

\begin{table}[htb]
\caption{Example of vocabularies with multi-reference}
\label{tab:multi_ref}
\centering
\fbox{
\begin{tabular}{lp{0.5\linewidth}}
\textbf{English} & \textbf{Chinese}\\
\midrule
\multirow{2}{*}{ferrous metal} & \begin{CJK*}{UTF8}{bkai}含鐵金屬(han tie jin shu) \end{CJK*}\\
& \begin{CJK*}{UTF8}{bkai} 黑色金屬(hei se jin shu)\end{CJK*}\\ \midrule
\multirow{3}{*}{chemical materials} & \begin{CJK*}{UTF8}{bkai}化學原料(hua xue yuan liao) \end{CJK*} \\
& \begin{CJK*}{UTF8}{bkai} 化學物品(hua xue wu pin)\end{CJK*}\\
& \begin{CJK*}{UTF8}{bkai} 化工材料(hua gong cai liao)\end{CJK*}\\ \midrule
\multirow{2}{*}{agricultural products} & \begin{CJK*}{UTF8}{bkai}農產品(nong chan pin) \end{CJK*}\\
& \begin{CJK*}{UTF8}{bkai}農業產品(nong ye chan pin) \end{CJK*}\\
\end{tabular}}
\end{table}

\subsubsection{Online Data}
As for the online dataset, parts of parallel data from the \textit{EMNLP 2017 SECOND CONFERENCE ON MACHINE TRANSLATION(WMT17)} Chinese-English for translation task are used due to the hardware resource limitation. The size of training data after filtering is around 1M which consists of about 252K sentence pairs from the \textit{News Commentary v12}\cite{Tiedemann} and 818k sentence pairs from \textit{CWMT Corpus}, including data from news, conversations, law documents, novels, etc. The vocabulary size of source language(English) is limited to 50k, which means if the vocabulary size is larger than 50k, only most common vocabularies will be chosen. For target language(Chinese), the size of vocabulary table is 6117. All the out-of-vocabulary words is mapped to a special token “UNK”. \textit{newstest2017} from \textit{WMT17} is chosen as test set. Statistics details of data is shown in Table \ref{tab:data_statistic}.

\begin{table*}
\caption{Data Statistics}
\label{tab:data_statistic}
\Centering
\resizebox{14cm}{!}{
\begin{tabular}{|cccccccccc|}
\hline
Data Source   & \multicolumn{5}{c}{Training Data} & \multicolumn{2}{c}{Development Data} & \multicolumn{2}{c|}{Testing Data} \\ \cline{2-10} 
\textbf{}              & \#OS     & \#PS   & SS\%   & LS\%   & \#AS & LS\%                   & \#AS                 & LS\%                  & \#AS               \\ \hline
\textit{Proposed Data} & 3057     & 2923   & 74.4\% & 25.6\% & 2.29 & 31.4\%                 & 2.42                 & 35\%                  & 2.31               \\
\textit{Online Data}   & 1070661  & 994849 & 92.7\% & 7.3\%  & 2.58 & 28.1\%                 & 2.29                 & 30.5\%                & 2.08               \\ \hline
\end{tabular}}
\begin{tablenotes}
\item \#OS: original number of sentences; \#PS: number of sentences after processing; SS\%: ratio of short sentences; LS\%: ratio of long sentences; AS\%: average number of segmented sequences per sentence\\
\end{tablenotes}
\end{table*}

\subsection{Experimental Setup}
Hyper-parameters set of training stage for coarse category network are: word embedding dimension as 128, number of LSTM layer for encoder as 4 with bidirectional LSTM on first layer and residual connection for the last two layer, number of normal LSTM layer decoder as 4, the number of hidden units for LSTM cell as 128, batch size as 256, gradient norm as 5.0. For fine category network, hyper-parameters are set: number of LSTM layer for encoder and decoder as 3 without bidirectional LSTM as well as residual connection while other hyper-parameters are the same as coarse category network.

SGD\cite{kiefer1952} optimizer is used with initial learning rate 1.0 as follows: for proposed data, train for 20K steps (around 1800 epochs); after 15K steps, learning rate will shrink to one tenth of the original every 1.5K step and the minimum learning rate will not less than 0.0001. For online data, train for 150K steps (around 16 epochs); after 120K steps, learning rate will shrink to one tenth of the original every 10K step and the minimum learning rate will not less than 0.0001. 

As our proposed dataset is relatively small, both the coarse category network and fine category network are fine tune on pretrained model. The pretrained model is self-trained model with simplify Chinese from online data. With the help of pretrained model, the model trained on proposed dataset would have a better initial state and may not fail into local optimization. 

Apart from the segmentation methods explained in Section \nameref{sec:sec_data_seg}, a more directed method for segmentation has also been applied to compared with the method in Section \nameref{sec:sec_data_seg}, which is segmenting long sentences only based on number of words. Roughly segmentation may also make sense to some extend. For every segmented short sequence in both source language and target language, a large part of the sequence are corresponding to each other. Also, for testing the impact of such mismatch brought by such method, we compared testing results of segmentation by number of words and by words as well as punctuation, when will be described in following section.

\subsection{Results}\label{sec:results}
In this section, experimental results will be shown. The quality of  translation are evaluated by 4-gram BLEU score\cite{bleu_1,bleu_2}. The compared models GNMT and Transformer are both proposed by Google in 2016 and 2017, respectively. For comparison networks, models are trained twice for experiments and reported BLEU scores will be the average results. Also, for evaluating the performance of networks on long sentences, the test set are split into test set of long sentences(test long data, with more than 50 words per sentence) and test set of short sentences(test short data).

As mentioned above, two segmentation methods are applied to proposed data. For the first one,  sentences with more than 50 words are segmented by words and punctuation while another method is segmenting long sentences by number of words as a comparison result. The testing result is shown in Table \ref{tab:diff_seg}. It can be seen from the table that segmentation with words and punctuation have a better performance then simply segmented by number of words. However, even though, segmented by number of words still have a good performance on proposed test set. It might because for most of provided data, the structures of sentences are more or less the same. So when long sentences are just simply segmented by fix number of words, the segmented sequences of both languages still have large parts of contents corresponding to each other. Also, it tells the proposed network has a certain degree of tolerance for such errors. 

\begin{table}[htb]
\caption{Evaluation results on proposed test set with different segmentation methods}
\label{tab:diff_seg}
\Centering
\resizebox{5cm}{!}{
\begin{tabular}{ccc}
\hline
                & SWP   & SN    \\\hline
full test data  & \textbf{62.218} & 60.096 \\
long test data  & \textbf{57.933} & 54.845 \\
short test data & \textbf{64.794} & 62.708 \\\hline
\end{tabular}}
\begin{tablenotes}
\item SWP: segmented by words and punctuation; SN: segmented by number of words; long test data: more than 50 words per sentence\\
\end{tablenotes}
\end{table}

Table \ref{tab:propose_gnmt_proposed} shows the predicting BLEU score on proposed test set. As the proposed network includes coarse category network and fine category network, prediction results from each parts of the proposed network are shown separately. Comparing results of coarse category network and GNMT, scores are better among all three dataset and is around 2.4 points better than GNMT on long test data while around 2 points better on short test data. After adding a fine category network following coarse category network, scores significantly better than GNMT, especially when fine tune the fine category network based on pretrained model, scores for both long test data and short test data are around 10 points better than that of GNMT.

\begin{table}[htb]
\caption{Comparison of results on proposed test set with GNMT}
\label{tab:propose_gnmt_proposed}
\Centering
\resizebox{7.8cm}{!}{
\begin{tabular}{cccccc}
\hline
\multirow{2}{*}{Data Types} & \multicolumn{3}{c}{Proposed Network} & {GNMT} \\ \cline{2-4} 
                  & C          & C+F        & C+F*          &    \\ \hline
full test data         & 54.883     & 56.517     & \textbf{62.218}  & 50.429    \\
long test data    & 51.171     & 53.635     & \textbf{57.933}  & 48.761    \\
short test data   & 57.402     & 59.199     & \textbf{64.794}  & 55.623    \\ \hline
\end{tabular}}
\begin{tablenotes}
\item C: coarse category network; F: fine category network; F*: fine category network based on pretrained model; long test data: more than 50 words per sentence\\
\end{tablenotes}
\end{table}

Table \ref{tab:propose_TRANSFORMER_proposed} shows the predicting BLEU score on proposed test set of proposed network and Transformer. The proposed network is 0.8 points better on long test sentences than that of Transformer. However, on short test data, the translation quality is 0.5 points better than that of proposed network. 
\begin{table}[htb]
\caption{Comparison of results on proposed test set with Transformer}
\label{tab:propose_TRANSFORMER_proposed}
\resizebox{8cm}{!}{
\begin{tabular}{ccc}
\hline
Data Types & {Proposed Network} & {Transformer} \\\midrule
full test data        &  \textbf{62.218}  & 59.381    \\
long test data   &  \textbf{57.933}  & 57.071    \\
short test data  &  64.794           & \textbf{65.242}    \\ \hline
\end{tabular}}
\begin{tablenotes}
\item long test data: more than 50 words per sentence\\
\end{tablenotes}
\end{table}

\begin{figure}[htb]
\centering
\includegraphics[height=5.5cm]{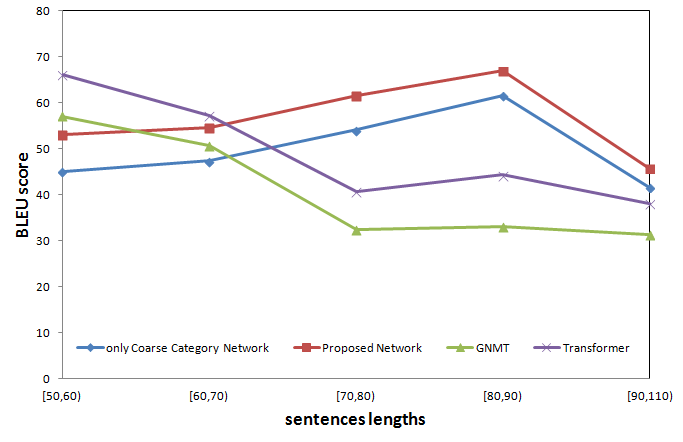}
\caption{Comparison of testing score grouped by sentences length \label{fig:test_in_length}}
\end{figure}

Targeting on long sentences, the proposed test data are divided into different groups according to sentences length and compared with GNMT and Transformer in each group. Based on Figure \ref{fig:test_in_length}, we observe that for sentences' length in range [50, 60), both GNMT and Transformer outperform the proposed network, especially for transformer, it is about 12 points better. However, the performance of GNMT goes down as the increasement of sentences length, especially when the sentences length is between 70 and 80. The Transformer almost works in the same way except when the sentences length is between 80 to 90, the score is slight increased. However, for the proposed network, as the length of sentences increased, the performance improved and until the length goes to 90, the BLEU score starts to go down. Also, when the sentences length larger than 70, the proposed network significantly outperforms GNMT and Transformer.

\begin{figure}[h]
\centering
\includegraphics[height=5.5cm]{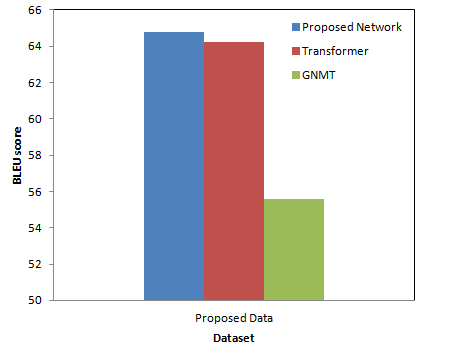}
\caption{Comparison on normal length data \label{fig:normal}}
\end{figure}

Besides, as shown in Figure \ref{fig:normal}, comparing the performance of normal length data in proposed data, the proposed network also outperform GNMT and slightly better than that of Transformer.

For online data from \textit{WMT17}, the results from proposed network and compared network are shown in Table \ref{tab:online_data_reslut}. Under same parameter setting and same size of dataset, the proposed network outperforms GNMT in all the test data set by 12 points, especially, the BLEU score of long test data win the highest score among three dataset in proposed data even though it is just around 0.2.

\begin{table}[htb]
\caption{Comparison of results on \textit{newstest2017}}
\label{tab:online_data_reslut}
\Centering
\resizebox{7cm}{!}{
\begin{tabular}{ccc}
\hline
{Data Types}     & {Proposed Network} & {GNMT} \\ \midrule
full test data   & \textbf{24.760}  & 13.204    \\
long test data   & \textbf{24.957}  & 12.517    \\
short test data  & \textbf{24.674}  & 11.426    \\ \hline
\end{tabular}}
\end{table}

\subsection{Analysis}
The proposed network is tested on the proposed test set as well as \textit{newstest2017} from \textit{WMT17} and the results are compared with GNMT and Transformer. 

On proposed test set, the proposed network significantly outperforms GNMT and Transformer on long sentences translation, especially when the sentences length is larger than 70(by around 24 points and 15 points, respectively). Also, scores of proposed network go up when the sentences length increases from 60 to 90, which tells the power of proposed network on translating long sentences. For the \textit{newstest2017}, the proposed network still outperforms GNMT by around 13 points. 

Also, the whole model is only trained on 2 GPUs(NVIDIA GeForce GTX 1080) with 8G on each and the size of training dataset is relatively small. If there is more hardware resources avaiable,  with a larger training dataset and a deeper network, the proposed network may have a better performance.

\section{Conclusions and Future Work}
In this paper, a hierarchical neural network based on sequence-to-sequence neural network for translation is proposed for improving the quality of long sentences translation. The proposed model is a cascaded of coarse category network which focus on learns semantic compositionality at short sequences level and a fine category network targeting on learns semantic compositionality as well as errors correction at sentences level. The model is compared with some currently state-of-the-art translation models GNMT as well as Transformer. It outperforms these models on long sentences translation and also have a comparable performance on normal length data. 

In the future, a better segmentation method will be explored to have better segmented short sequences. We will try to introduce monolingual data into our model so as to break the limitation of rare parallel English-Chinese data pairs. 

\bibliographystyle{paper}
\bibliography{paper}

\end{document}